\documentclass[letterpaper, 10 pt, conference]{ieeeconf}  

\IEEEoverridecommandlockouts                              
\usepackage{amsthm}
\usepackage{amsmath,amsfonts}
\usepackage{algorithmic}
\usepackage{array}
\usepackage[caption=false,font=normalsize,labelfont=sf,textfont=sf]{subfig}
\usepackage{textcomp}
\usepackage{stfloats}
\usepackage{url}
\usepackage{verbatim}
\usepackage{graphicx}
\usepackage{cite}
\usepackage{nameref}
\usepackage{hyperref}
\usepackage{epsfig} 
\usepackage{mathtools}
\usepackage{amssymb}  
\usepackage[ruled, vlined, linesnumbered]{algorithm2e}
\usepackage{todonotes}
\usepackage{xcolor}
\usepackage{graphicx}
\usepackage{hhline}

\usepackage{multirow}
\usepackage{mathrsfs}
\usepackage{ctable} 
\usepackage{booktabs,ragged2e}
\usepackage[english]{babel}
\usepackage{gensymb}
\usepackage{hhline}
\usepackage{hyperref}
\usepackage{amssymb} 
\usepackage[T1]{fontenc} 
\usepackage{mathtools} 

\usepackage{kotex}

\theoremstyle{definition}

\newtheorem*{running}{Running Example}

\def\equationautorefname~#1\null{(#1)\null}
\def\sectionautorefname~#1\null{Section #1\null}
\def\subsectionautorefname~#1\null{Section #1\null}
\def\algorithmautorefname~#1\null{Algorithm #1\null}
\def\theoremautorefname~#1\null{Theorem #1\null}
\def\lemmaautorefname~#1\null{Lemma #1\null}
\def\remarkautorefname~#1\null{Remark #1\null}
\def\definitionautorefname~#1\null{Definition #1\null}

\usepackage{tikz}

\newcommand\copyrighttext{%
  \footnotesize \textcopyright 2025 IEEE.  Personal use of this material is permitted.  Permission from IEEE must be obtained for all other uses, in any current or future media, including reprinting/republishing this material for advertising or promotional purposes, creating new collective works, for resale or redistribution to servers or lists, or reuse of any copyrighted component of this work in other works.}
\newcommand\copyrightnotice{%
\begin{tikzpicture}[remember picture,overlay]
\node[anchor=south,yshift=10pt] at (current page.south) {\fbox{\parbox{\dimexpr\textwidth-\fboxsep-\fboxrule\relax}{\copyrighttext}}};
\end{tikzpicture}%
}

\begin{document}

\title{\LARGE \bf
Enhancing Feature Tracking Reliability for Visual Navigation\\using Real-Time Safety Filter
}
\author{Dabin Kim$^{*}$, Inkyu Jang$^{*}$, Youngsoo Han, Sunwoo Hwang, and H. Jin Kim
\thanks{$^{*}$ The first two authors contributed equally to this work.}
\thanks{This work was supported by the National Research Foundation of Korea (NRF) grant funded by the Korea government (MSIT) (RS-2024-00436984). The work by Inkyu Jang was partially supported by Basic Science Research Program through the National Research Foundation of Korea (NRF) funded by the Ministry of Education (RS-2024-00407121).}
\thanks{The authors are with the Department of Aerospace Engineering, Seoul National University, Seoul 08826, South Korea (e-mail: dabin404@snu.ac.kr, janginkyu.larr@gmail.com, swsw0411@snu.ac.kr, cat7945@snu.ac.kr, hjinkim@snu.ac.kr, corresponding author: H. Jin Kim).}
}

\maketitle
\copyrightnotice
\thispagestyle{empty}
\pagestyle{empty}
\vspace{-3.5mm}

\begin{abstract}
Vision sensors are extensively used for localizing a robot's pose, particularly in environments where global localization tools such as GPS or motion capture systems are unavailable. In many visual navigation systems, localization is achieved by detecting and tracking visual features or landmarks, which provide information about the sensor's relative pose. For reliable feature tracking and accurate pose estimation, it is crucial to maintain visibility of a sufficient number of features.
This requirement can sometimes conflict with the robot's overall task objective. In this paper, we approach it as a constrained control problem. By leveraging the invariance properties of visibility constraints within the robot's kinematic model, we propose a real-time safety filter based on quadratic programming. This filter takes a reference velocity command as input and produces a modified velocity that minimally deviates from the reference while ensuring the information score from the currently visible features remains above a user-specified threshold.
Numerical simulations demonstrate that the proposed safety filter preserves the invariance condition and ensures the visibility of more features than the required minimum. We also validated its real-world performance by integrating it into a visual simultaneous localization and mapping (SLAM) algorithm, where it maintained high estimation quality in challenging environments, outperforming a simple tracking controller.


\end{abstract}

\section{Introduction}

Vision sensors are widely used for self-localization in mobile robots. Visual Odometry (VO) and Visual Simultaneous Localization and Mapping (V-SLAM) are extensively researched in both the computer vision and robotics fields. State-of-the-art visual pose estimation algorithms, such as ORB-SLAM \cite{mur2017orb} and VINS-Mono \cite{qin2018vins}, have proven to be highly effective.

While most research has traditionally focused on improving the accuracy and robustness of visual estimation using available image data, recent studies have begun examining the impact of image data quality on the performance of vision-based localization algorithms. The quality of image data is often influenced by the camera's trajectory. For example, when a camera follows a trajectory that captures texture-less surfaces (e.g., plain walls), the accuracy of VO may decrease due to the lack of sufficient visual features. This challenge has sparked a growing interest in perception-aware planning and control, where the camera trajectory is adjusted to ensure high-quality visual data for reliable localization. This idea has led to several studies that integrate visual estimation considerations into advanced motion planning and control strategies.

\begin{figure}[t]
    \centering
    \subfloat[]{
    \includegraphics[width=0.9\linewidth]{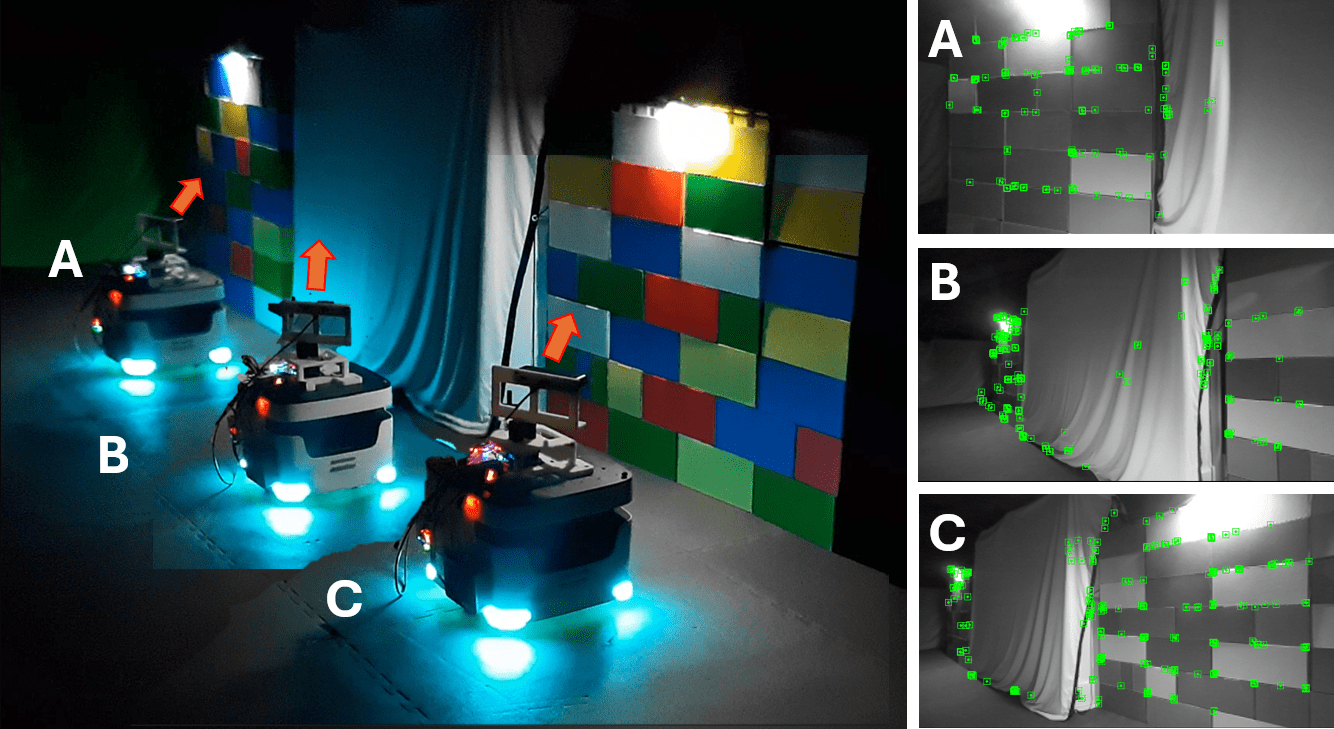}}
    \\ \vspace{-2mm}
    \subfloat[]{
    \includegraphics[width=0.9\linewidth]{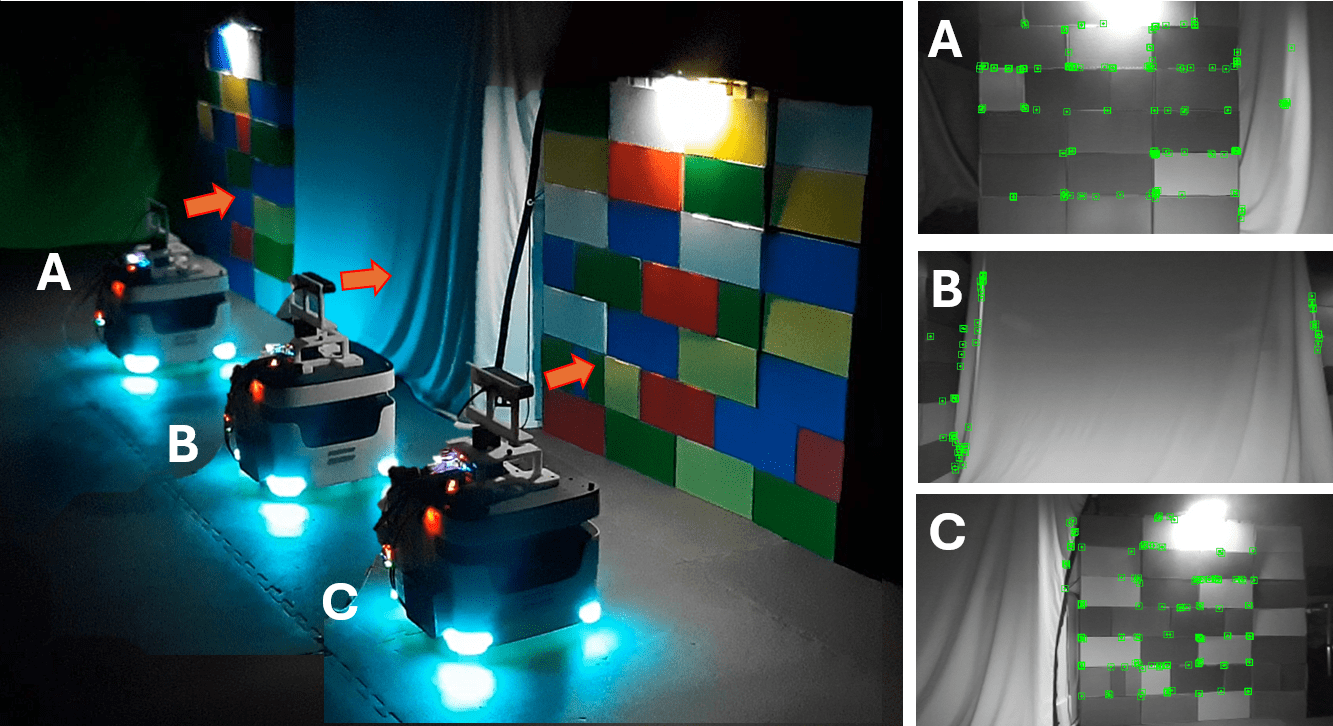}}
    \caption{The result of experiments with (a) the proposed safety filter and (b) the baseline controller. For each experiment, the robots at three timestamps (A,B,C) are visualized along with their corresponding on-board images, with features visualized in ORB-SLAM2. The detected features are represented with green dots. The proposed safety filter adaptively  adjusts the control input to maintain sufficient tracking features, as demonstrated by the camera heading (arrow) and onboard image at timestamp B. 
     In contrast, the baseline controller struggles with texture-poor surfaces, where fewer features are trackable. A detailed explanation and analysis are provided in \autoref{sec:eval}.}   
    \label{fig:exp_proposed}
\end{figure}

Belief-space planning and optimal control methods \cite{indelman2015planning} \cite{van2011lqg}, which rely on explicit state estimation and uncertainty modeling, are often unsuitable for real-time control in modern V-SLAM systems due to the high computational load of tasks such as  bundle adjustment involving hundreds of landmarks. To address this limitation, recent research has explored incorporating information directly extracted from image inputs, such as feature points, into the planning and control loop. These algorithms typically frame the problem as a multi-objective optimization, where perception requirements become an additional objective for the controller.
However, this multi-objective approach introduces significant challenges. Balancing path optimality with maintaining sufficient visual information is difficult, and comparing Pareto-optimal solutions to find the best trade-off adds complexity. Additionally, prioritizing visual information can often degrade trajectory quality, potentially leading to numerical instability when the multiple objectives are in conflict.


In this paper, rather than adopting a multi-objective optimization approach, we address the problem from a safety-critical control perspective. Instead of prioritizing the optimization of the estimation algorithm's performance, our focus is on ensuring sufficient visual information for stable localization by framing it as a constraint in the optimization process. This approach is based on the observation that once a sufficient amount of information is obtained, adding more does not significantly enhance estimation performance.


Building on the theory of safety filters which ensure the forward invariance of a safety set, we developed a visibility maintenance condition designed to guarantee a sufficient number of visual features within an image, despite a limited field of view. From these visibility maintenance conditions, we formulate a quadratic program (QP)-based safety filter that guarantees feasibility and supports real-time computation. Unlike previous works on visibility-constrained control such as \cite{kim2023visibility}, which focused on maintaining fixed targets within a limited field of view, our approach accommodates newly observed or lost features dynamically. To the best of the authors' knowledge, this is the first attempt to develop a safety filter for perception-aware control in visual navigation that guarantees feasibility. We conduct numerical evaluations to validate that the designed safety filter can maintain a user-defined threshold for the minimum number of visual features. Additionally, experimental results are presented to demonstrate the efficacy of the proposed method using real-world data and a visual estimator.

\section{Related Work}
\subsection{Planning and Control for Reliable Visual Navigation}

The importance of acquiring rich keypoints in visual navigation has been well recognized since the early days of visual SLAM research \cite{davison2002simultaneous}. With advancements in path planning and control, there has been a growing focus on integrating visual navigation capabilities into these algorithms.

Belief Space Planning (BSP) addresses state uncertainty by modeling it as a probability distribution, allowing objectives like collision avoidance to be treated as chance constraints. For instance, estimators with explicit uncertainty representation, such as the Extended Kalman Filter (EKF), have been combined with Rapidly-exploring Random Trees (RRT) to manage uncertainty in path planning \cite{bry2011rapidly}. 
In vision-based systems, computationally intensive techniques such as bundle adjustment are used to estimate future states. Integrating these methods into the core of planning and control algorithms is often impractical for real-time control due to their high computational cost \cite{achtelik2014motion} \cite{indelman2015planning}. To mitigate this computational burden, metrics like the Fisher Information Matrix \cite{zhang2019beyond} or the observability Gramian \cite{hausman2017observability} are employed as proxies for explicit uncertainty computation and are used as objective functions in optimal control algorithms.

In contrast, vision-based heuristics are designed to assess visual navigation quality efficiently, often derived directly from pixel measurements. For instance, point visibility can be constrained in controller design \cite{falanga2018pampc} \cite{kim2023visibility}, effective for fixed landmarks but unsuitable for dynamic  feature tracking. Metrics like the number of visible features \cite{kim2021topology} and co-visible features \cite{chen2024apace} approximate visual estimation quality and integrate into trajectory optimization. Our work aligns closely with this, operating directly in image space. However, instead of balancing multiple objectives in trajectory optimization, we use a safety-critical control method that formulates reliable feature tracking as a constrained control problem.
The resulting QP-based safety filter provides safe control with feasibility guarantees under mild assumptions, offering a more interpretable and numerically efficient real-time control method compared to multi-objective non-convex trajectory optimization approaches.

\subsection{Safety-Critical Control with Perception Objective}

The growing focus on autonomous navigation has increased the need for safety with formal guarantees like collision avoidance. Safety-critical control techniques including Hamilton-Jacobi reachability and control barrier functions (CBFs) have been developed to address these needs. These techniques now increasingly integrate perception systems for broader applications.

To handle state uncertainty, measurement-robust CBFs \cite{cosner2021measurement} and observer-controller co-designs \cite{agrawal2022safe} address safety under estimation errors. In vision-based systems, the challenge of high-dimensional image inputs is tackled by learning-based CBFs trained on RGB-D images to avoid collisions with arbitrarily shaped objects \cite{abdi2023safe}, and BarrierNet \cite{xiao2023barriernet}, which uses differentiable CBFs for tasks like end-to-end driving. CBFs have also been designed for environmental representations such as point clouds from depth cameras \cite{de2024point} and neural radiance fields \cite{tong2023enforcing} for collision avoidance.

Beyond collision avoidance, safety-critical control can also be used to maintain the visibility of points of interest, which is essential in applications like visual servoing \cite{zheng2019toward} \cite{wei2024diffocclusion} and teleoperation \cite{kim2023visibility}, ensuring that targets remain within the camera's field of view. We extend this concept by developing a safety filter for reliable visual navigation, enforcing a sufficient number of visible image features.



\section{Real-Time Constraint Satisfaction using QP-Based Safety Filter}


Assume a continuous-time time invariant nonlinear system model
\begin{equation}
    \dot{x} = f(x) + g(x) u,
\end{equation}
where $x \in \mathbb{R}^n$ is the state, $u\in U \subseteq \mathbb{R}^m$ is the input. The set $U$ is assumed to be a convex polytope in the $\mathbb{R}^m$ space, i.e., there exist a matrix $A_u$ and a vector $b_u$ with appropriate sizes such that $U = \{u\in \mathbb{R}^m : A_u u \leq b_u\}$.
We assume that for every reachable state $x$ in the state space, there exists an input $u\in U$ (possibly as a function of $x$) such that $f(x) + g(x) u = 0$. This $u$ is called the \textit{stopping input} that brings the system to an instantaneous complete stop. 
Relaxing this sudden stop assumption is an interesting future research direction which will be discussed in the conclusion section.

Given a set of allowed states $C \subseteq \mathbb{R}^{n}$, we want to generate a feedback law $u(t,x) \in U$ as a function of time $t$ and state $x$, such that the system permanently resides in $C$, i.e., $C$ is forward invariant.
Suppose there exist a finite number of continuously differentiable functions $h_i:\mathbb{R}^n \rightarrow \mathbb{R}$, $\forall i \in I$, where $I$ is the index set, such that $C = \{x\in \mathbb{R}^n : h_i(x) \geq 0\}$, and $\partial_x h_i(x) \neq 0$ if $h_i(x) = 0$.
Then, Nagumo's theorem \cite[Section 4.2]{blanchini2008set} tells that if $u$ is chosen such that
\begin{equation} \label{eq:nagumo thm}
    h_i(x) = 0 \rightarrow \partial_x h_i (x) \cdot \left(f(x) + g(x) u\right) \geq 0, \; \forall i \in I,
\end{equation}
then the set $C$ will be rendered forward invariant.

In order to build a feedback control strategy that best tracks the given reference input while satisfying \autoref{eq:nagumo thm} and avoiding discontinuity, one can let $u(t,x)$ be the optimal solution to the following quadratic program:
\begin{equation} \label{eq:velocity filter}
\begin{aligned}
    \min_{u \in \mathbb{R}^m} \; & (u - u_\mathrm{ref}(t,x))^\top R(u - u_\mathrm{ref}(t,x)) \\
    \mathrm{s.t.} \; & \dot{h}_i(x, u) \geq -\alpha_i h_i(x), \; \forall i \in I \\
    & A_u u \leq b_u,
\end{aligned}
\end{equation}
where $R\in \mathbb{R}^{m\times m}$ is a symmetric positive definite weight matrix, $\dot{h}_i(x,u) = \partial_x h_i(x) \cdot (f(x) + g(x)u)$ is the time derivative of $h_i(x)$ given state $x$ and input $u$, $\alpha_i$-s are positive reals.
Note that given a strictly positive definite $R$, \autoref{eq:velocity filter} admits a unique solution which satisfies \autoref{eq:nagumo thm}. Additionally, the optimization is feasible given $x \in C$, since the stopping input is one feasible solution to the optimization.
Thus, \autoref{eq:velocity filter} can be called a \textit{safety filter} \cite{wabersich2023data, hsu2023safety} in the sense that it selects the input from the set of safe inputs $U_{f} = \{u : \text{$u$ is a feasible solution to \autoref{eq:velocity filter}}\}$ that minimally deviates from the reference.
This approach can be regarded as a special case of CBF-based quadratic programs (CBF-QP) \cite{ames2016control} where the instantaneous brakability assumption allows to employ multiple CBFs in a set-intersection manner without suffering from \textit{leaking corner} issues \cite{lee2019removing}.

\section{Visibility Maintenance using Safety Filter}


\subsection{Differential Kinematics}
Let the robot's configuration space be $Q \subseteq \mathbb{R}^n$. We model the robot's differential kinematics as follows:
\begin{equation}
    \dot{q} = J(q)v, \label{eq:kinematics}
\end{equation}
where $J(q):Q\rightarrow \mathbb{R}^{n\times m}$ is the Jacobian matrix which we assume continuous with respect to $q$, $v \in V \subseteq \mathbb{R}^m$ is the input, and $V$ is the input constraint set which is assumed to be a convex polytope such that $V \ni 0$ (i.e., it satisfies the instantaneous brakability condition).

We consider a vision sensor attached rigidly to the robot frame (so that its pose $T \in SE(3)$ is given as a continuously differentiable function of the robot's configuration $q$), which is capable of detecting positions of point landmarks within its region of detection. To elaborate, let $p$ be the position of a landmark seen in the vision sensor frame. The condition for the landmark to be seen from the sensor can be written as $\rho(p) \geq 0$, where $\rho:\mathbb{R}^3 \rightarrow \mathbb{R}^d$ is the continuous and differentiable constraint function.

The motion of the sensor frame can be written using its body linear and angular velocities (i.e., twist). If we assume the landmarks are fixed to the world frame (i.e., stationary), the landmark position with respect to the sensor $p$ is controllable by the body twist and follows the dynamics
\begin{equation} \label{eq: p dynamics}
    \dot{p} = -\omega_s \times p - v_s,
\end{equation}
where $\omega_s \in \mathbb{R}^3$ and $v_s\in \mathbb{R}^3$ are the sensor's body angular and linear velocities, and $\times$ is the vector cross product in 3D.
Since $T$ is a continuously differentiable function of $q$, we can then write $\omega_s$ and $v_s$ as a linear function of the control input $v$ as follows:
\begin{equation} \label{eq: input transform}
    \omega_s = J_\omega(q) v, \;\; v_s = J_v(q) v,
\end{equation}
where $J_\omega$ and $J_v$ are matrix-valued continuous functions of $q$ with appropriate sizes.

\begin{figure}
    \centering
    \hspace{1mm}
    \subfloat[]{
        \includegraphics[width=0.45\linewidth]{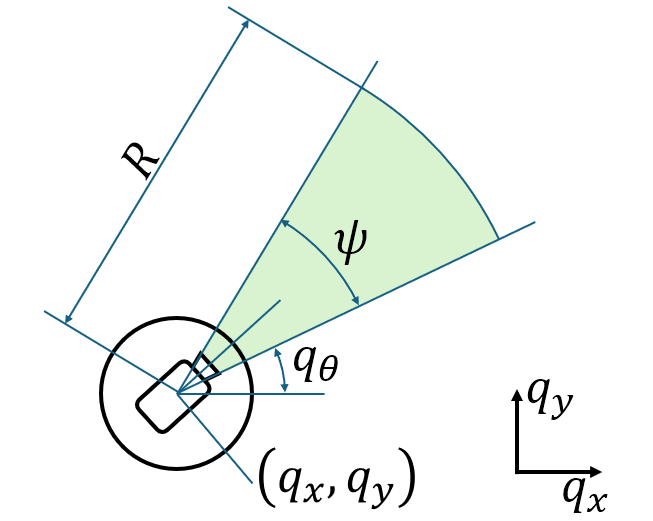}
        
    }
    \subfloat[]{
        \includegraphics[width=0.4\linewidth]{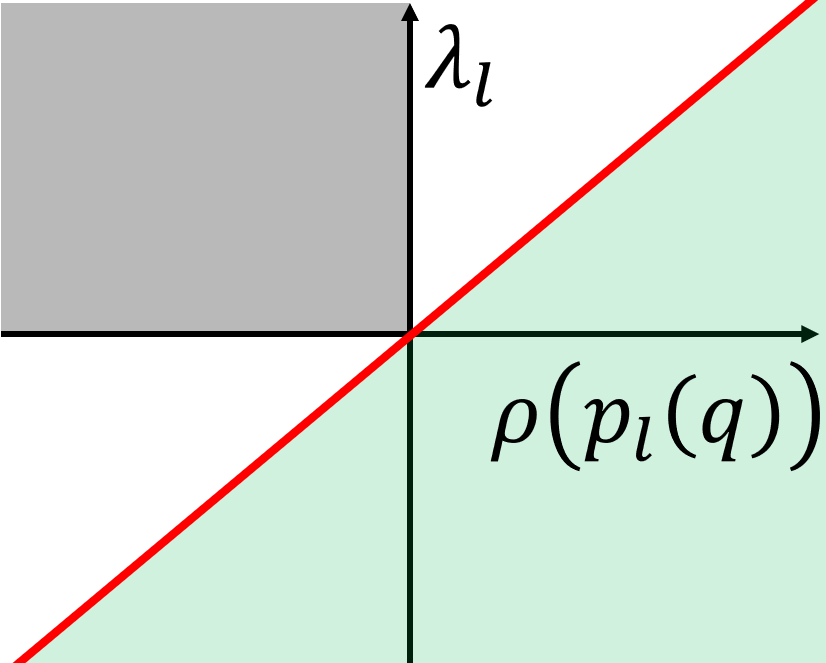}
    }
    \caption{(a) The robot configuration and the onboard camera's field of view for the running example. The camera is mounted on the two dimensional ground robot, captures any landmark within its field of view, represented by the light green region. The field of view is defined by the angle of view, $\psi$, and the sensing range, $R$.
    (b) A graphical illustration of equivalence between \autoref{eq: lambda nonnegativity} and \autoref{eq: lambda-rho}. The gray region shows where \autoref{eq: lambda nonnegativity} is violated. For a fixed $\mu \in [0,1]$, the feasible region of \autoref{eq: lambda-rho} forms a half-space with the origin on its boundary. The union of all possible half-spaces aligns with the feasible set of \autoref{eq: lambda nonnegativity}, represented by the white region.}
    \label{fig: running example}
\end{figure}

\begin{running}[Ground robot]
    In this section, as a running example, we consider the following ground robot (\autoref{fig: running example}) with a camera attached on it:
    \begin{equation}
        \dot{q} = \frac{d}{dt}\begin{bmatrix}
            q_x \\ q_y \\ q_\theta
        \end{bmatrix} = v = \begin{bmatrix}
            v_x \\ v_y \\ v_\theta
        \end{bmatrix},
    \end{equation}
    where $q_x$, $q_y$ are the horizontal and vertical positions of the robot base, $q_\theta$ is the orientation (rotation angle with respect to the $x$ axis) of the camera, $v_x$, $v_y$, $v_\theta$ are control inputs.
    It is straightforward to find that the landmark point kinematics \autoref{eq: p dynamics} can be written as
    \begin{equation}
        \dot{p} = \frac{d}{dt}\begin{bmatrix}
            p_x \\ p_y
        \end{bmatrix} = \begin{bmatrix}
            -\cos q_\theta & -\sin q_\theta & p_y \\
            \sin q_\theta & -\cos q_\theta & -p_x
        \end{bmatrix} \begin{bmatrix}
            v_x \\ v_y \\ v_\theta
        \end{bmatrix}.
    \end{equation}
    Here, $p_x$ and $p_y$ are the horizontal and vertical positions of the landmark seen from the camera frame, respectively.
\end{running}

\subsection{Visibility Maintenance} \label{sec:vis_maintenance}
Let $L$ be the set of $N$ landmarks in three-dimensional space, where the position of each landmark $l \in L$ relative to the vision sensor is $p_l(q)$. A landmark $l$ is visible if $\rho(p_l(q)) \geq 0$, and the set of visible landmarks is denoted as $L(q) := \{l \in L : \rho(p_l(q)) \geq 0\}$.

\begin{running}[Ground robot, continued]
    We assume that the onboard camera can detect landmarks within the field of view as shown in \autoref{fig: running example} (a). Given the field of view, the $\rho$ function can be defined as follows:
    \begin{equation}
        \rho(p) = \begin{bmatrix}
            [\sin \psi/2, \cos \psi/2]^\top p \\
            [\sin \psi/2, -\cos \psi/2]^\top p \\
            R - \lVert p\rVert_2
        \end{bmatrix}
    \end{equation}
    where $\psi$ is the angle of view, $R$ is the sensing range. In this example, we let $R = 1$, $\psi = 1\text{ rad}$.
\end{running}

Given all above, we want the robot's motion to satisfy  the following constraint:
\begin{equation}
    w(q) := \sum_{l \in L(q)} w_l \geq W,
\end{equation}
where $w_{(\cdot)} \geq 0$ are the weights, {$w(q)$ represents the overall landmark tracking quality score, $w_{l}$ denotes the contribution of landmark $l$ to this quality, and $W$ is the minimum score the robot is required to achieve throughout the mission.}
For example, if the goal is to maintain visibility to at least $M$ landmarks, we can let $w_l = 1$ for all $l\in L$ (so that $w(q)$ counts the number of currently visible landmarks), and $W$ any positive number between $M-1$ and $M$.

Unfortunately, since $w(q)$ is not everywhere differentiable, it is impossible to construct a QP-based safety filter like \autoref{eq:velocity filter}. Instead, following the idea of \cite{jang2023invariance}, we introduce an auxiliary state variable $\lambda = [\lambda_1, \cdots, \lambda_N]^\top \in \mathbb{R}^N$ and a \textit{smoothened} version of score constraint
\begin{equation}
    W \leq \hat{w}(q, \lambda) = \sum_{l \in L} \lambda_l w_l \leq w(q).
\end{equation}
The first inequality $W \leq \hat{w}$ is continuously differentiable with respect to $\lambda$ and has no dependency on $q$. For the second inequality $\hat{w}\leq w$ to hold, we require
\begin{equation}
    \lambda_l \leq \mathbf{1}_{l\in L(q)}(q)
\end{equation}
where $\mathbf{1}_{\phi(\xi)}(\xi)$ is the indicator function that returns $1$ if the statement $\phi(\xi)$ is true and $0$ otherwise.
This can be rewritten as the intersection of two constraints for every $l \in L$: 
\begin{align}
    \lambda_l &\leq 1, \label{eq: lambda <= 1} \\
    \lambda_l > 0 \; &\rightarrow \; \rho(p_l(q)) \geq 0 \label{eq: lambda nonnegativity}.
\end{align}
The first inequality \autoref{eq: lambda <= 1} is continuously differentiable, and Eq. \autoref{eq: lambda nonnegativity} is equivalent to  
\begin{equation}
    \lambda_l \leq 0 \,\, \vee \,\, \rho(p_l(q)) \geq 0,
\end{equation}
which holds if and only if
\begin{equation} \label{eq: lambda-rho}
    \exists \mu_l \in [0,1], \quad -\mu_l \lambda_l + (1 - \mu_l) \rho(p_l(q)) \geq 0.
\end{equation}
The proof for this claim is straightforward and hence omitted in this paper. A graphical explanation to this can be found in \autoref{fig: running example} (b).
To combine everything into a continuously differentiable setting, we let $\mu = [\mu_1, \cdots, \mu_N]^\top \in \mathbb{R}^N$ be another auxiliary state variable of the system. The auxiliary state $\lambda$ and $\mu$ are \textit{controlled} through the dynamics
\begin{equation} \label{eq: mu lambda dynamics}
    \dot{\lambda} = v_\lambda, \quad \dot{\mu} = v_\mu,
\end{equation}
where $v_\lambda, v_\mu \in \mathbb{R}^{N}$ are virtual inputs that are numerically integrated to obtain the actual $\lambda$ and $\mu$ values.

\subsection{QP-based Safety Filter}

In summary, we have the following set of state constraints
\begin{equation} \label{eq: final state constraints}
\begin{aligned}
    h_1(x) &= \sum_{l \in L}\lambda_l w_l - W \geq 0, \\
    h_{2,l}(x) &= 1 - \lambda_l \geq 0, \;\forall l \in L, \\
    h_{3,l}(x) &= -\mu_l \lambda_l + (1 - \mu_l)\rho(p_l(q)) \geq 0, \;\forall l \in L, \\
    h_{4,l}(x) &= \mu_l \geq 0, \; \forall l \in L, \\
    h_{5,l}(x) &= 1 - \mu_l \geq 0, \; \forall l \in L, \\
    h_6(x) &= c(q),
\end{aligned}
\end{equation}
with $x = (q, \lambda, \mu) \in \mathbb{R}^{n + 2N}$ being the augmented state. The constraints $h_1$ through $h_{5,l}$ are derived from \autoref{sec:vis_maintenance}, and $h_6$ is the collision avoidance constraint encoded as a function of robot state $c(q)$ which we assume continuously differentiable. For example, one can model the robot as a sphere and let $c(q) = s(q) - r$, where $s(q)$ is the (signed) distance from the robot to the nearest obstacle, $r>0$ is the radius of the robot.

These state constraints are continuously differentiable with respect to the augmented state $x$ and a safety filter in the form \autoref{eq:velocity filter} can be implemented:
\begin{equation} \label{eq: final safety filter}
\begin{aligned}
    \min_{u \in U} \quad & (u - u_\mathrm{ref}(t,x))^\top R (u - u_\mathrm{ref}(t,x)) \\
    \mathrm{s.t.} \quad & \dot{h}_{(\cdot)}(x,u) \geq -\alpha_{(\cdot)} \cdot h_{(\cdot)}(x)
\end{aligned}
\end{equation}
where the inequality constraint should be satisfied for all $h_{(\cdot)}$ functions in \autoref{eq: final state constraints}.
Here, we let $u=(v, v_\lambda, v_\mu) \in U = V \times \mathbb{R}^N \times \mathbb{R}^N \subseteq \mathbb{R}^{m + 2N}$ be the augmented input and $u_\mathrm{ref}(t, x) = (v_\mathrm{ref}(t, q), 0, 0)$, $R = \mathrm{blkdiag}(R_q, k_\lambda \mathrm{1}_N, k_\mu \mathrm{1}_N)$ where $v_\mathrm{ref}$ is the reference input given from the higher-level decision maker (e.g., manual control from a human operator or a motion planning algorithm), $R_q \in \mathbb{R}^{m\times m}$ is the symmetric positive definite input cost matrix, $k_\lambda$ and $k_\mu$ are positive (usually very small) weights to ensure the safety filter has a unique solution, $\mathrm{1}_N$ is the identity matrix of size $N\times N$.

The time derivatives of $h_{(\cdot)}$ can be expressed as a linear function of $u$, thus the safety filter \autoref{eq: final safety filter} is a QP.
The $\lambda$ and $\mu$ values should be calculated by integrating \autoref{eq: mu lambda dynamics}.
In case $p_l(q)$ and its derivative are not straightforward to directly compute, one can evaluate it in an indirect manner by forward integrating to get the following initial value ODE, which follows directly from \autoref{eq: p dynamics} and \autoref{eq: input transform}:
\begin{equation}
\begin{aligned}
    \dot{p}(t) &= -J_\omega(q(t)) v(t) \times p(t) - J_v(q(t)) v(t), \\ p(t_0) &= p_l(q(t_0))
\end{aligned}
\end{equation}
where $p_l(q(t_0))$ is the position of the landmark in sensor frames when initially observed at time $t_0$, and $p$ is a shorthand for $p_l \circ q$.
It can be seen that if $h_{(\cdot)}(x) \geq 0$, the optimization \autoref{eq: final safety filter} is feasible since $u=0$ makes $\dot{h}_{(\cdot)} = 0$ and thus is one feasible solution.

\begin{running}(Ground robot, continued)
    In the running example, the mission of the robot is to track the reference input, while maintaining visibility to at least 5 landmarks. Thus, we let $W = 4.5$, $w_l=1$.
    We place $N=30$ landmarks at random positions near the origin. The reference input is given as
    \begin{equation}
        v_\mathrm{ref}(t,q) = \begin{bmatrix}
            -\sin t + 2 (\cos t - q_x) \\
            \cos t + 2 (\sin t - q_y) \\
            0
        \end{bmatrix}
    \end{equation}
    which makes the robot to track a circular trajectory around the origin at speed $1$.
    In this example, we ignore collision avoidance constraints ($h_6$ in \autoref{eq: final state constraints}).
    Other parameter values are set as follows: $\alpha_{(\cdot)} = 1$, $V = [-2, 2]\times [-2, 2] \times [-1, 1]$, $R = \mathrm{diag}(1, 1, 0.001)$, $k_\lambda = k_\mu = 0.001$. The third component of $R$ corresponding to $v_\theta$ is set to a small number to allow the camera to rotate accordingly to keep the landmarks in sight.
\end{running}

\subsection{Initialization and Handling Limited Observability}

To take advantage of the nonnegativity-preserving property of the safety filter, it is very important to initialize the safety filter with a valid augmented state $x$ so that $h_{(\cdot)} \geq 0$.
Suppose the robot starts at an initial condition $q(t_0)$ such that $w(q(t_0)) \geq W$ and $c(q(t_0)) \geq 0$. This means that the robot has view to a sufficient amount of information from the landmarks and is at a collision-free position. Then, one can easily find that the initialization
$\lambda_l(t_0) = \mathbf{1}_{l\in L(q(t_0))}(l)$, $\mu_l(t_0) = \mathbf{1}_{l\notin L(q(t_0))}(l)$ is a valid choice that $h_{(\cdot)}(x) \geq 0$.

In real deployment, the landmark positions $p_l$ are typically not \textit{a priori} known and the robot can observe only the currently visible landmark positions at discrete times.
Suppose that the observations occur at times $t_0 < t_1 < \cdots$. Whenever $t=t_i$ for an integer $i \geq 0$, the robot newly observes $p_l$ such that $l\in L_i = L(q(t_i))$. During the interval $t \in \left[t_i, t_{i+1}\right)$ for every $i \in \{0,1,\cdots\}$, we require the robot to run the safety filter with $L_i$ instead of $L$, with re-initialization $\lambda_l(t_i) = \mathbf{1}_{l\in L_i}(l) = 1$, $\mu_l(t_0) = \mathbf{1}_{l\notin L_i}(l) = 0$. This will introduce a sudden jump in the $h_{(\cdot)}$ values at time $t=t_i$, however, here we show that the jumps preserve nonnegativity of $h_{(\cdot)}$ values.
Let $x_i^- = (q_i, \lambda_i^-, \mu_i^-)$ and $x_i^+ = (q_i, \lambda_i^+, \mu_i^+)$ be the augmented states before and after the jump at time $t_i$ ($i \in \{1,2,\cdots\}$), respectively. Note that the robot configuration $q$ does not jump. Suppose $h_{(\cdot)}(x_i^-) \geq 0$.
Firstly, $h_1$ value always jump \textit{upwards}, i.e., $h_1(x_i^+) \geq h_1(x_i^-) \geq 0$, since $h_1(x_i^-) \leq w(q_i) - W$ (the nonnegativity of $h_{(\cdot)}(x_i^-)$ ensures this), and $h_1(x_i^+) = \sum_{l\in L_i} w_l - W = w(q_i) - W$.
Next, it is straightforward to find the values of $h_{2,l}$ through $h_{5,l}$ are nonnegative after the update.
Finally, $h_6$ only depends on $q$ which does not jump, therefore $h_6(x_i^+)=h_6(x_i^-)\geq 0$.

In summary, if the initial conditions are such that $w(q(t_0)) \geq W$ and $c(q(t_0)) \geq 0$, then the robot will start from a feasible condition that satisfies \autoref{eq: final state constraints}. 
Moreover, for every $i \in \{0,1,\cdots\}$, the safety filter will ensure the constraint satisfaction throughout $t \in [t_i, t_{i+1})$, and the jump at $t=t_{i+1}$ will set the $h_{(\cdot)}$ value to a nonnegative one, after which the safety filter can resume from a feasible initial condition.

\begin{figure}
    \centering
    \vspace{2mm}
    \includegraphics[width=0.97\linewidth]{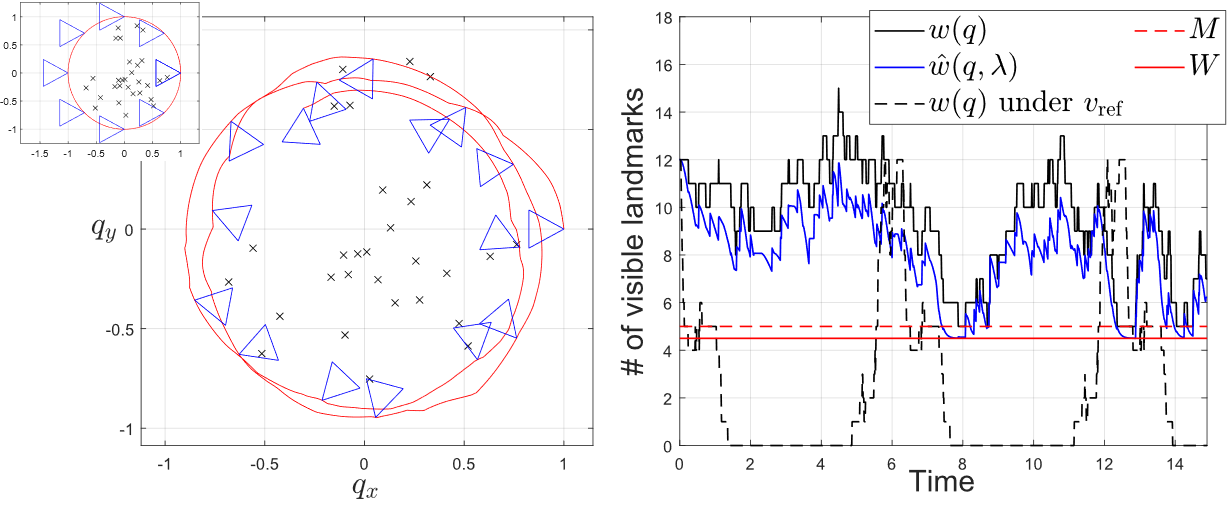}
    \caption{Simulation result for the running example. (left) The resulting trajectory of the robot. The position trajectory ($q_x, q_y$) is depicted as red curve, the camera poses are drawn using blue triangles. It can be seen that the robot takes a path that differs from the reference (shown in the left top corner) to maintain visibility to at least 5 landmarks. (right) The time history of $w(q)$, $\hat{w}(q,\lambda)$ values. The relation $W \leq \hat{w}(q,\lambda) \leq w(q)$ holds throughout the simulation.}
    \label{fig: running example simulation result}
\end{figure}
\begin{running}[Ground robot, simulation result]
    With the abovementioned setting and initialization, we simulated the safety filter using MATLAB. The results are depicted in \autoref{fig: running example simulation result}. In the results, it can be clearly seen that the $\hat{w}$ value always lower bounds the number of visible landmarks $w$ and lower bounded by $W$, resulting in $w(q) \geq 5$ throughout the simulation. The robot takes the path that provides visibility to at least 5 landmarks, while minimally deviating from the reference input given. Note that the re-initialization introduces sudden jumps in the $\hat{w}(q, \lambda)$ values, but only in a way that the state constraints are satisfied.
\end{running}





\section{Evaluation} \label{sec:eval}



\begin{figure}
    \centering
    \includegraphics[width=0.9\linewidth]{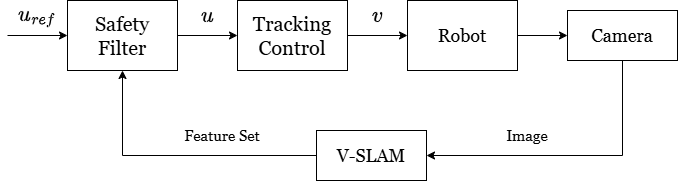}
    \caption{The diagram illustrated for the control structure for vision-based robot with the proposed safety filter.}
    \label{fig:scheme}
\end{figure}

\begin{figure}
    \centering
    \subfloat[]{
    \includegraphics[width=0.48\linewidth]{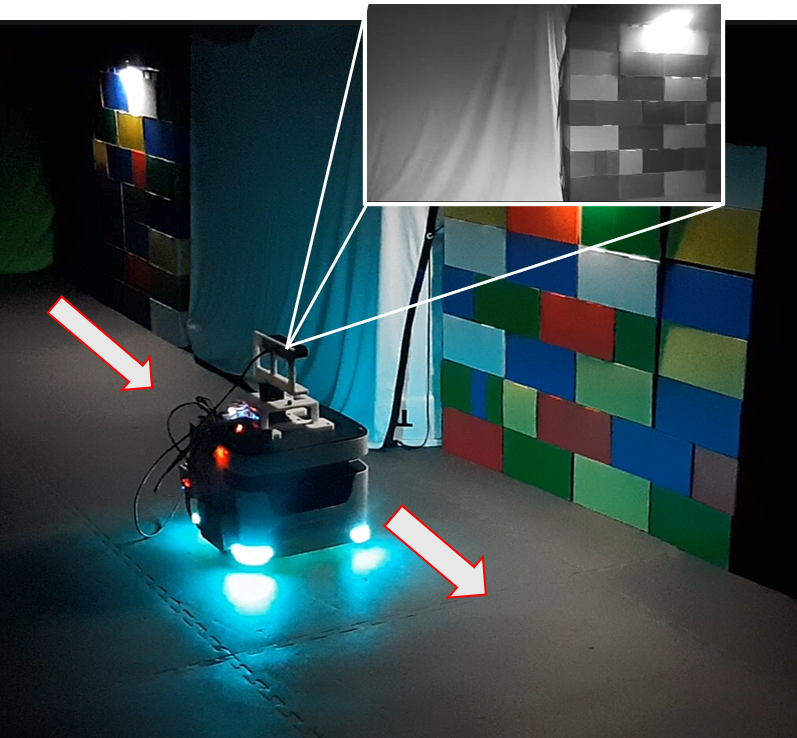}
    }
    \subfloat[]{
    \includegraphics[width=0.41\linewidth]{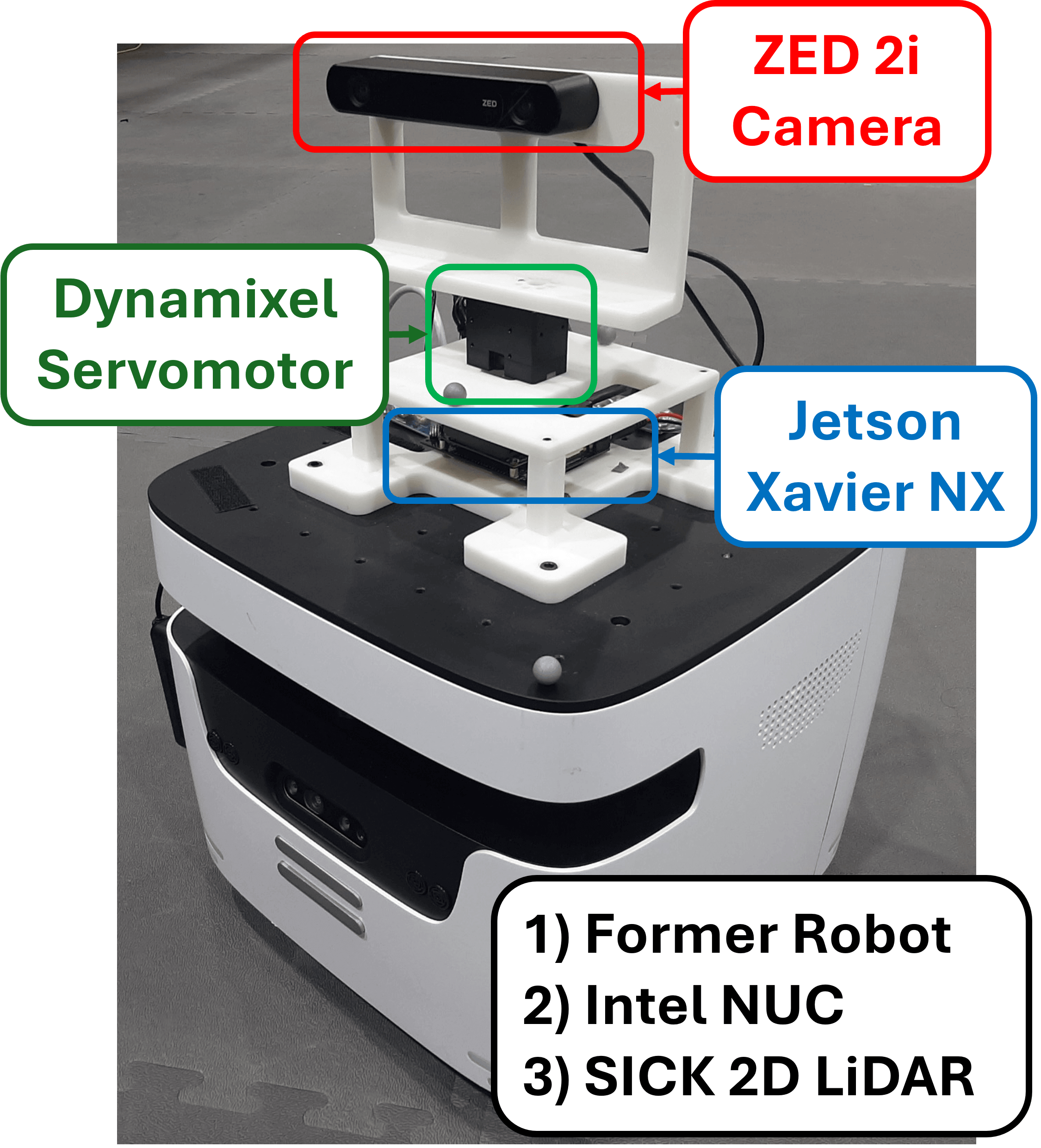}
    }
    \caption{(a) The experiment scenario where the robot inspects the wall moving through the texture-poor region and (b) the hardware configuration used for the hardware experiment. }
    \label{fig:hardware}
\end{figure}

The proposed safety filter is validated on real hardware in an experimental setup in a vision-challenging environment.

\subsection{Mission Objective and Hardware Configuration}

The mission is to inspect a wall using a stereo camera, which includes both texture-rich and texture-poor regions, as shown in Fig. \ref{fig:hardware} (a). The objective is to move along the wall while keeping the camera oriented perpendicular to the surface, ensuring an optimal view of the inspection area. However, the texture-poor region poses a challenge for visual navigation.

The hardware setup, shown in Fig. \ref{fig:hardware} (b), includes a differential wheeled robot (ROAS Inc.) with a ZED2i stereo camera mounted on a servomotor-equipped rig, allowing rotation aligned with the robot's axis. An NVIDIA Jetson Xavier NX handles is paired with the stereo camera for real-time processing, and a SICK 2D LiDAR is used for obstacle detection.
All system components are connected via Ethernet communication.

\subsection{Implementation Details}
The control structure with the proposed safety filter and hardware is shown in Fig. \ref{fig:scheme}. The robot state and the velocity are defined as $q = [q_{x},q_{y},\theta_{r}, \theta_{m}]^{T}$ and $v=[v_{r},\omega_{r},\omega_{m}]^{T}$, where $\theta_{r}$, $v_{r}$ and $\omega_{r}$ are heading angle, velocity and angular velocity of the base robot and $\theta_{m}$ and $\omega_{m}$ are the heading angle and angular velocity of the servomotor. The Jacobian matrix $J$ is $\begin{bsmallmatrix}
\cos \theta_{r} & 0 & 0 \\
\sin \theta_{r} & 0 & 0 \\
0 & 1 & 0 \\ 
0 & 0 & 1
\end{bsmallmatrix}$,
and the heading angle of the camera is $\theta_{c}=\theta_{r}+\theta_{m}$.

The reference linear velocity and angular velocity (\(v_r\), \(\omega_r\)) for the base robot, and the reference angular velocity (\(\omega_{s,r}\)) for the servomotor are given to the safety filter. \(v_r\) is set as positive constant value and \(\omega_{r}\) to zero when the heading \(\theta_{r}\) 
 is set to the direction parallel to the wall. The servomotor is controlled by a PID controller to regulate the camera to be aligned with the wall’s normal vector.

For visual navigation, we employ ORB-SLAM2 \cite{mur2017orb} with stereo images. Each extracted feature is measured as $m=[m_{u},m_{v},m_{d}]^{T}=[f_{x} \frac{p_{x}}{p_{z}} + c_{x}, f_{y} \frac{p_{y}}{p_{z}} + c_{y}, d]^{T}$, where \( (f_{x}, f_{y}) \) represents the focal lengths, \( (c_{x}, c_{y}) \) denotes the principal point, and \( d \) is the depth measurement. The visibility condition $\rho$ is defined as $m_{u} \in [0, I_{w}],m_{v}\in [0,I_{h}]$, and $m_{d}\in [r_{min},r_{max}]$ where $I_{w}$ and $I_{h}$ are the width and height of the image, and $r_{min}$ and $r_{max}$ are the depth thresholds for valid detection.  
Since the number of inequality constraints linearly increases to the number of features, we limit the features passed to the safety filter to $N_{max}=50$ to meet the real-time requirement for solving the QP problem \eqref{eq: final safety filter} by sampling $N_{max}$ number of features from the total feature set. The signed distance function from $h_{6}$ in \eqref{eq: final state constraints} is obtained from 2D LiDAR data.

\subsection{Results}

\begin{figure}
    \centering
    \vspace{-2mm}
    \subfloat[]{
        \includegraphics[width=0.5\linewidth]{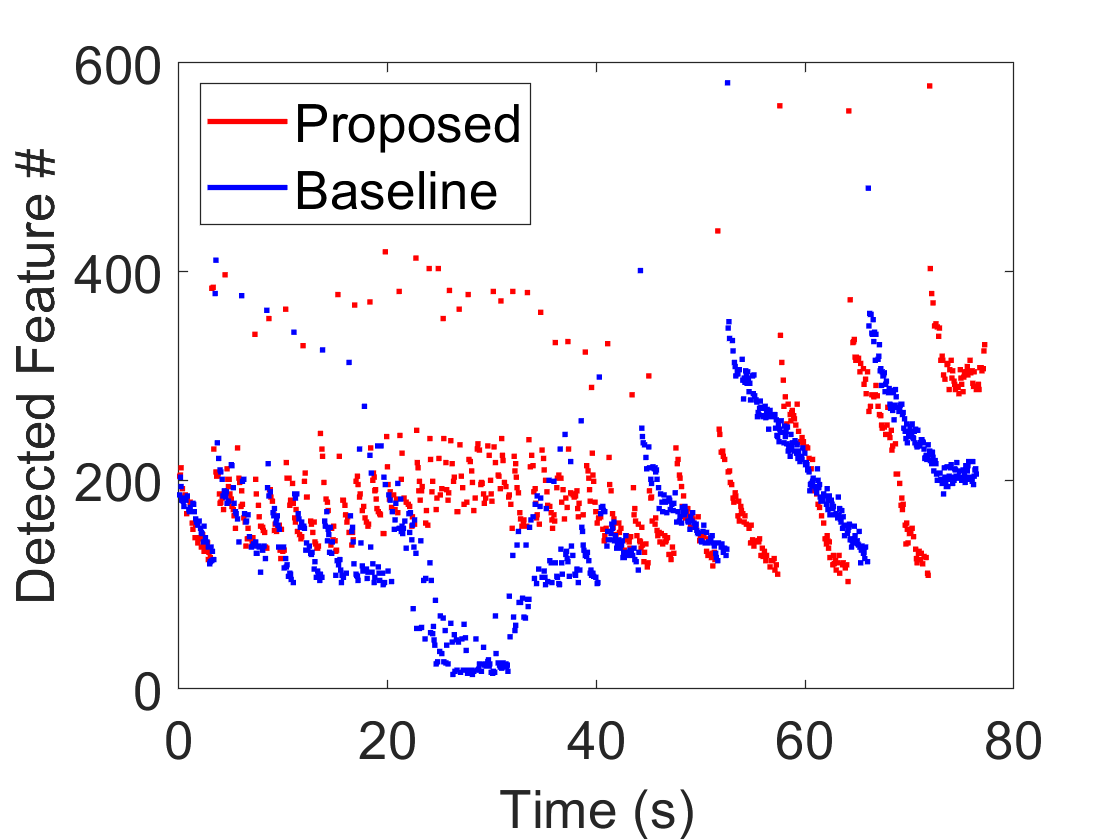}}
    \subfloat[]{
        \includegraphics[width=0.5\linewidth]{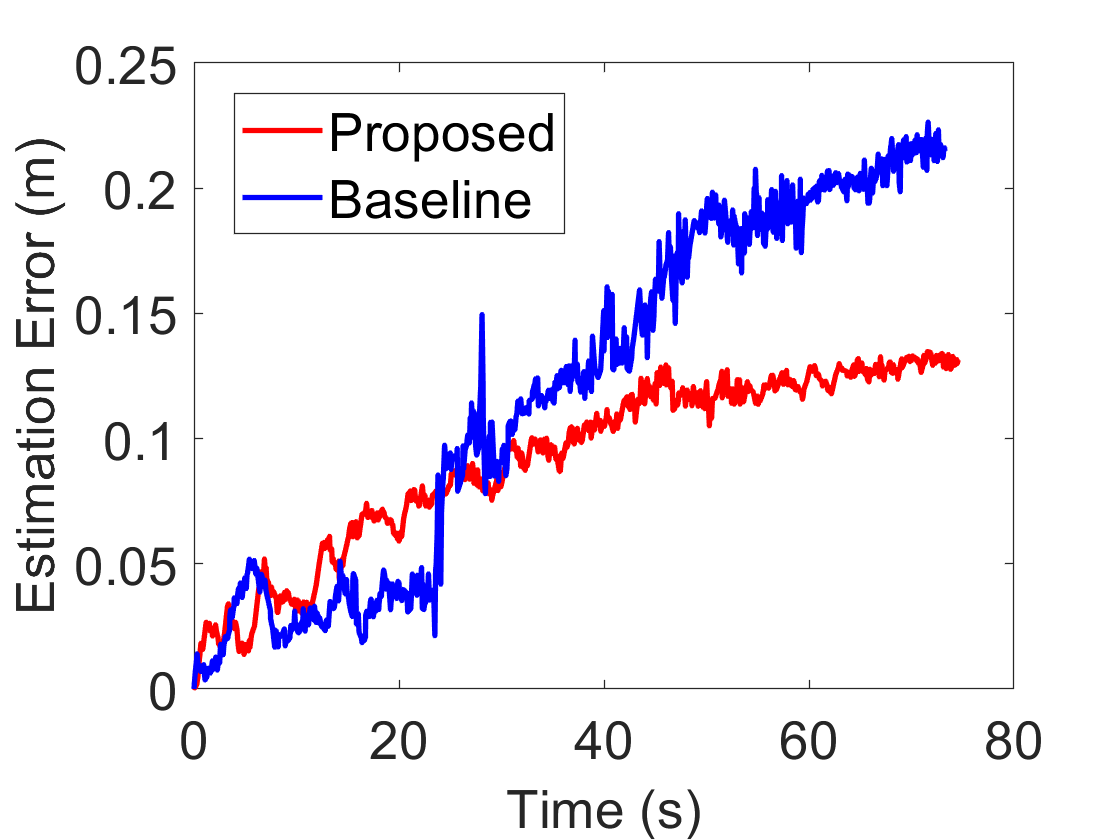}}
    \caption{Comparison of the result of the visual SLAM algorithm in  between the proposed safety filter (red) and the baseline controller (blue). (a) The number of tracked features and (b) the state estimation error acquired from the ORB-SLAM2 algorithm.}    
    \label{fig:exp_estimation_result}
\end{figure}

In the baseline algorithm, the reference velocity is applied directly to the robot's tracking controller, with the servomotor keeping the camera perpendicular to the wall. As shown in the onboard images of Fig. \ref{fig:exp_proposed} (a), the robot moves through a texture-poor region, causing a shart drop in visible features and an increase in estimation error (Fig. \ref{fig:exp_estimation_result}). 

In contrast, the proposed safety filter  adjusts the control input to maintain a sufficient number of visible features, even when passing through feature-poor regions. It rotates the camera toward feature-rich regions to ensure continuous tracking, returning to the orientation close to the desired orientation when the robot reaches a  feature-rich area. As a result, a sufficient number of features are kept tracked as in Fig. \ref{fig:exp_estimation_result} (a). We observe that, in this scenario, our algorithm successfully maintains a sufficient number of visible features, implying it can be used along with the visual front-end with real-world data.

\section{Conclusion}

Despite the widespread use of vision sensors in mobile robot navigation, they remain susceptible to challenges like low-texture environments and poor lighting. While researchers have developed trajectory planners and controllers to maximize visual information alongside control objectives, these methods often suffer from parameter sensitivity and numerical instability. 

To address this, we proposed a safety-critical control approach using a QP-based safety filter to maintain sufficient visual features in an image with a novel safety condition, enabling real-time implementation.
Experiments show it integrates seamlessly with visual odometry or SLAM, reducing the risk of catastrophic failures in visual estimation.

Future work will relax assumptions like the need for a stopping input to extend applicability to higher-relative-degree systems. We will also tackle  practical challenges such as occlusion, which can abruptly reduce observed landmarks and disrupt the invariance condition. Additionally, we plan to explore integrating other visual estimation quality proxies, such as informativeness \cite{zhang2019beyond} and system excitation \cite{hausman2017observability}, into the proposed safety filter.


\bibliographystyle{./bibtex/IEEEtran}
\bibliography{./bibtex/IEEEabrv, ./bibtex/mybibfile}

\addtolength{\textheight}{-12cm}   

\end{document}